\documentclass[runningheads]{llncs}
\usepackage[T1]{fontenc}
\usepackage{graphicx}
\usepackage{subfigure}
\usepackage{booktabs}
\usepackage{makecell}
\usepackage{tabularx}
\usepackage{multicol}
\usepackage{multirow}
\usepackage{amsmath}
\usepackage{xcolor}
\usepackage{amssymb}
\usepackage {amsmath}
\usepackage{bbding}
\usepackage{url}
\usepackage{diagbox}
\usepackage{graphicx,verbatim}
\begin{document}
\title{Bridging Single Distortion Artifacts and Multifactorial Clinical Quality: Few-shot Biparametric MRI Quality Assessment via Distortion-trained Prototypical Networks}
\titlerunning{Few-shot Biparametric MRI Quality Assessment}
%

\author{
Yucheng Tang\inst{1} \and 
Alexander Ng\inst{2} \and 
Wen Yan\inst{1} \and 
Natasha Thorley\inst{3} \and 
Pawel Rajwa\inst{2} \and 
Yipei Wang\inst{1} \and 
Aqua Asif\inst{4,2} \and 
Clare Allen\inst{5} \and 
Louise Dickinson\inst{5} \and 
Francesco Giganti\inst{2,5} \and 
Shonit Punwani\inst{5,6} \and 
Daniel C. Alexander\inst{7,8} \and 
Veeru Kasivisvanathan\inst{2,9} \and 
Yipeng Hu\inst{1} 
}

\authorrunning{Tang et al.}

\institute{
UCL Hawkes Institute and Department of Medical Physics and Biomedical Engineering,
University College London, UK; \email{yucheng.tang.24@ucl.ac.uk}
\and
Division of Surgery and Interventional Science,
University College London, UK
\and
Centre for Medical Imaging,
University College London, UK
\and
British Urology Researchers in Surgical Training (BURST), UK
\and
Department of Radiology,
University College London Hospitals NHS Foundation Trust, UK
\and
Centre of Medical Imaging, Division of Medicine,
University College London, UK
\and
Centre for Medical Image Computing,
University College London, UK
\and
Department of Computer Science,
University College London, UK
\and
Department of Urology,
University College London Hospitals NHS Foundation Trust, UK
}
  
\maketitle              
\begin{abstract}
Clinical prostate multi-parametric MRI relies heavily on high-quality diffusion-weighted imaging (DWI), yet reading DWI is frequently compromised by geometric distortion, often caused by rectal air.
Assessing quality via the PI-QUAL scoring system is an emerging clinical standard, but it is subjective, time-consuming and suffers from a class imbalance where low-quality cases are diverse and relatively scarce. 
Using the PRIME clinical trial as an example, there are $6\%$ images with PI-QUAL scores lower than 4, $87\%$ of DWI issues are due to distortion. Many of the other clinical quality issues are under-represented. To address this common dual-scarcity of annotated clinical data, we propose a few-shot biparametric prototypical network for automated image quality assessment (IQA). Our framework utilizes a dual-branch 3D ResNet to fuse T2-weighted and DWI features, providing anatomical context to distinguish true morphology from distortion. To handle real-world heterogeneity, we introduce feature-wise linear modulation (FiLM) and a gradient reversal layer (GRL) to align feature distributions conditioned on varying b-values while suppressing acquisition-related biases. We demonstrate that a model meta-trained solely on comparatively objective, readily obtainable distortion labels can effectively adapt to predicting complex, multi-factorial clinical quality scores such as PI-QUAL using only five representative samples. Experimental results on two datasets show that our method significantly outperforms few-shot learning baselines for this challenging IQA task, offering a practically feasible and data-efficient solution for standardizing prostate MRI quality control in clinical workflows.
Code is available at \url{https://anonymous.4open.science/r/Proto-FM-IQA-2627}.


\end{abstract}
\section{Introduction}
Multi-parametric and biparametric magnetic resonance imaging (MRI) is becoming a noninvasive modality in prostate cancer management, which typically includes T2-weighted imaging (T2WI) and diffusion-weighted imaging (DWI)~\cite{ng2025biparametric}. DWI is highly sensitive to clinically significant cancer, but inherently susceptible to artifacts related to echo-planar imaging acquisition and susceptibility effects, which cause geometric distortion~\cite{donato2014geometric}, lesion displacement and deformation, significantly compromising the reliability of lesion localization, zonal assessment and its subsequent multiparametric analysis. 

To standardize the quality assessment of multi-parametric prostate MRI, the PI-QUAL scoring system~\cite{giganti2020prostate} has been developed. Rather than assessing a single sequence in isolation, PI-QUAL is a multifactorial metric based on the combined diagnostic quality of all acquired sequences, primarily T2WI and DWI. For example, a PI-QUAL score of $< 4$ indicates insufficient diagnostic quality to reliably rule out clinically significant lesions. Notably, a low PI-QUAL score does not exclusively imply a poor DWI. However, when severe geometric distortion compromises the diagnostic interpretability of DWI, it often serves as the primary factor for downgrading the overall PI-QUAL score.
In current clinical practice, PI-QUAL scoring relies on subjective assessment by experienced radiologists, which is time-consuming with substantial inter-observer variation~\cite{van2026real}. 
 
Automating image quality assessment (IQA) for prostate MRI faces a dual-scarcity constraint~\cite{oksuz2019automatic}. First, although non-optimal scans are common~\cite{giganti2023global}, low-quality cases are inherently rare in optimized clinical trials (e.g., PRIME) and well-established centers, leading to severe class imbalance. Second, at the quality-type level, the few available low-quality DWI scans are typically dominated by susceptibility-induced geometric distortions, providing insufficient examples of other clinically relevant issues.
Strategies like resampling on scarce data risk overfitting rather than learning generalizable distortion features~\cite{shorten2019survey}. Standard augmentations fail to simulate the complex geometric distortions of echo-planar imaging~\cite{jezzard1995correction}. Furthermore, although k-space simulation is promising, it often requires auxiliary data (e.g., field maps)~\cite{andersson2003correct} that are unavailable in retrospective cohorts, limiting performance on real-world clinical data.

We propose a two-stage few-shot learning strategy. In the first stage, we curate a larger-scale distortion dataset to explicitly learn distortion-sensitive features under an episodic prototypical meta-learning. In the second stage, such learned model is directly applied to out-of-domain datasets or the PI-QUAL scoring system by re-computing prototypes for the new task distribution. By utilizing only five clinical samples per class for adaptation, the network successfully generalizes from specific physical deformations (here, DWI distortion) to the complex clinical requirements of diagnostic-grade prostate MRI.

Our contributions are as follows. First, we propose a novel few-shot biparametric prototypical network for automated prostate IQA. 
Second, we introduce a dual-branch architecture that processes T2WI and DWI in parallel, where T2WI serves as a reference to effectively differentiate true morphology from geometric distortion in DWI. Third, we incorporate feature-wise linear modulation (FiLM) and a gradient reversal layer (GRL) to align feature distributions conditioned on varying b-values while explicitly suppressing acquisition-related domain biases.
Extensive experiments on two labelled datasets demonstrate superior performance compared to state-of-the-art few-shot learning approaches. 

\section{Method}

\begin{figure*}[t]
    \centering
    \includegraphics[width=\textwidth]{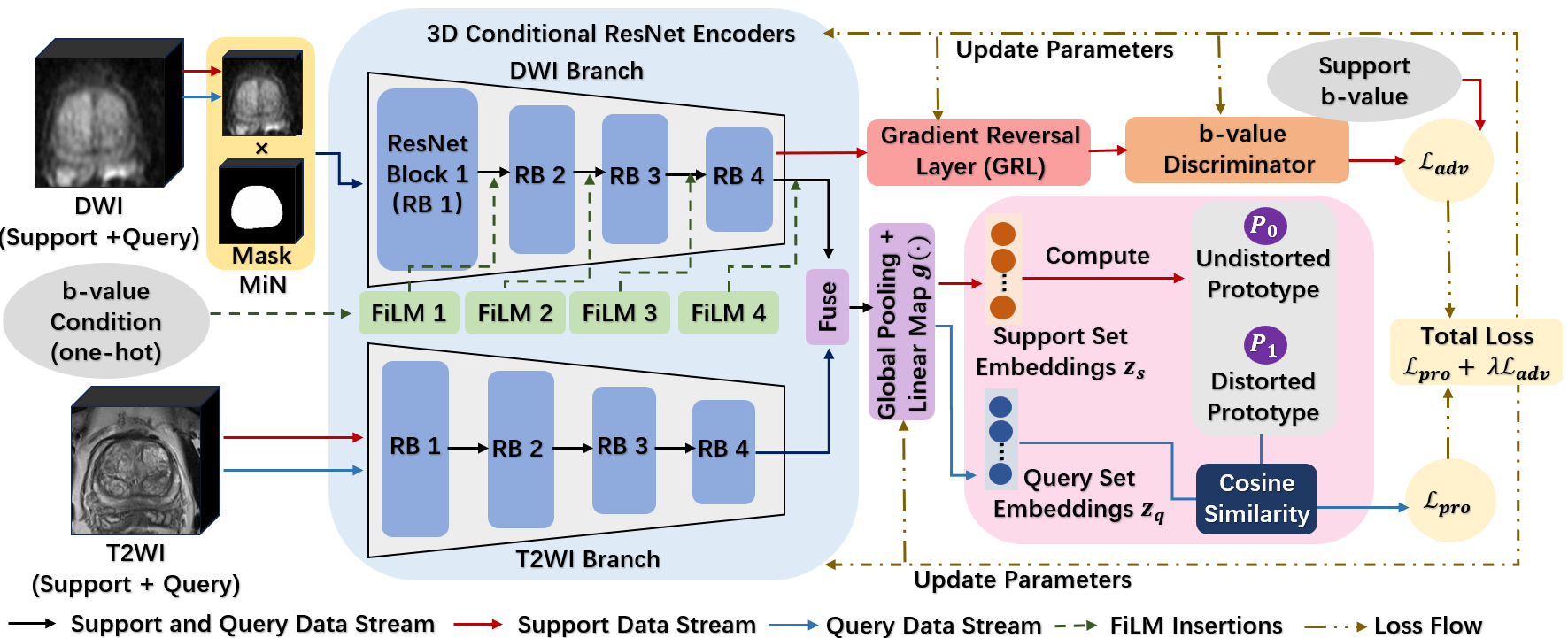}
    \caption{Overview of the proposed method. The network utilizes a dual-branch encoder to process DWI and T2WI sequences. FiLM layers and the GRL are applied specifically to the DWI branch to handle b-value variations. The features are fused to compute prototypes and perform query IQA.}
    \label{fig:main}
\end{figure*}

\subsection{Distortion Feature Learning via Meta-training}
To address the scarcity of low-quality training data, we first utilize a larger dataset with distortion labels to learn distortion-sensitive feature representations.
We formulate this stage as an $N$-way $K$-shot $Q$-query episodic few-shot learning task~\cite{vinyals2016matching}, where $N = 2$ denotes the number of quality classes (distorted vs. undistorted), $K$ and $Q$ represent the number of labeled samples per class in the support and query set, respectively.
In each training episode, the model is presented with a support set $\mathcal{S}=\{(x_i^s,y_i^s,b_i^s)\}_{i=1}^{N\times K}$ and a query set $\mathcal{Q}=\{(x_j^q,y_j^q,b_j^q)\}_{j=1}^{N\times Q}$. Here, $x_i^s$ and $x_j^q$ are the input image pairs for the support and query instances, respectively; $y_i^s$ and $y_j^q$ denote their binary distortion labels (e.g., $0$ for undistorted, $1$ for distorted); and $b_i^s$ and $b_j^q$ represent their one-hot encoded b-value conditions.
As shown in Fig.~\ref{fig:main}, the 3D conditional ResNet encoders take DWI and T2WI pairs as input.
The DWI branch with FiLM extracts distortion-sensitive features. These features are then passed through a GRL and a b-value discriminator to facilitate adversarial learning. The T2WI branch provides anatomical context to distinguish true morphological variations from distortions.
The features from the four ResNet blocks of both branches are fused by an attention fusion module~\cite{woo2018cbam}.
The fused representation is then subjected to a global pooling and a linear mapping $g(\cdot)$ to generate an embedding $z$ for each DWI and T2WI pair. Subsequently, the embeddings of the support set are used for prototype construction, and the embeddings of the query set are compared against these prototypes to compute the prototypical loss.

\textbf{Masked Instance Normalization.}
To reduce the influence of background intensity variations and encourage prototypes to focus on meaningful tissue regions, we employ masked instance normalization (MiN). Given an input image $x$, we first generate the foreground using a binary mask generated by mean intensity thresholding: $m = \mathbb{I}~(x > \bar{x})$, where $\bar{x}$ is the global mean intensity of $x$ and $\mathbb{I}$ is the indicator function. We then compute the within-mask mean 
and variance 
to normalize the foreground voxels to zero-mean and unit-variance, while strictly setting all voxels outside the mask to zero.

\textbf{Combined FiLM and GRL Modules.}
We incorporate both FiLM and a GRL into the DWI branch, where FiLM adapts feature distributions to the varying b-values used during the scan, while the GRL suppresses acquisition-related domain biases.
Let $F$ be the input feature map and $v$ be its corresponding acquisition b-value. Since the dataset only contains a limited number of specific b-values, we first map $v$ to a one-hot encoded vector $b \in \{0, 1\}^B$ with $\|b\|_1 = 1$, where $B$ is the total number of discrete b-values. Next, we project $b$ into a dense embedding vector $e_b \in \mathbb{R}^d$ using a learnable embedding table $\mathbf{E} \in \mathbb{R}^{B \times d}$, where $d$ matches the channel dimension of $F$. This embedding is projected via a linear layer to generate channel-wise scaling and shifting parameters $[\gamma(b),\beta(b)] = W_E \cdot e_b + b_E$, where $W_E$ and $b_E$ are learnable weights and biases. These parameters modulate the feature map $F$ as $F^{'} = \gamma(b)\odot F + \beta(b)$, where $\odot$ denotes element-wise multiplication. 
To encourage the learned IQA representations to be invariant to the conditioning b-value, which is always readily available for DWI data, we introduce an adversarial mechanism~\cite{tzeng2017adversarial} with a GRL.
While $F^{'}$ is forwarded to the fusion module for the main task, a copy of it is passed through the GRL to a b-value discriminator. The GRL acts as an identity transform during the forward pass but reverses the gradient sign during backpropagation by multiplying with a scalar $-\alpha$ ($\alpha > 0$). 
The following b-value discriminator $D$ predicts the b-value $b$ from the features as $\hat{b} = Softmax(W_D\cdot F^{'} + b_D)$.

\textbf{Loss Function.}
The total training loss comprises prototypical and adversarial losses. For prototypical loss, we first compute the undistorted and distorted prototypes $P_0$ and $P_1$ by averaging the embeddings of the support samples as $P_c = \frac{1}{|\mathcal{S}_c|} \sum_{(x_i, y_i) \in \mathcal{S}_c} \frac{z_i}{\|z_i\|_2}$, where $c = \{0,1\}$ and $\mathcal{S}_c$ is the subset of support samples belonging to class $c$. Then, for each query sample $x_j^q$, we compute the probability of it belonging to class $c$ as $\mathcal{P}(y_j^q=c~|~x_j^q) = \frac{\exp(\tau \cdot \cos(z_j^q, P_c))}{\sum_{c'=0,1} \exp(\tau \cdot \cos(z_j^q, P_{c'}))}$, where $\cos(\mathbf{a}, \mathbf{b}) = \frac{\mathbf{a} \cdot \mathbf{b}}{\|\mathbf{a}\| \|\mathbf{b}\|}$ denotes cosine similarity and $\tau$ is a scaling factor.
The prototypical loss is defined as $\mathcal{L}_{pro}=CE(y_j^q, \mathcal{P}(y_j^q=c~|~x_j^q))$, where $CE(\cdot)$ represents the cross-entropy loss. The adversarial loss is defined as $\mathcal{L}_{adv}= CE(b, \hat{b})$. This objective forces the encoder to generate b-value-invariant representations, ensuring that the final embedding captures quality-related semantics rather than acquisition metadata.
The total loss is calculated as the weighted sum: $\mathcal{L} = \mathcal{L}_{pro} + \lambda \mathcal{L}_{adv}$, where all learnable parameters in the meta-training phase are updated by minimizing this objective.

\subsection{Meta-testing and Adaptation}
During both the meta-testing (in-domain) and adaptation (out-of-domain or cross-task) phases, the model parameters are frozen. Consistent with the meta-training strategy, the evaluation on a given dataset $\mathcal{D}_{eva}$ is formulated as a few-shot task. 
Specifically, the support set $\mathcal{S}_{eva}$ is constructed using $K$ representative samples per class from $\mathcal{D}_{eva}$. 
These samples are chosen to cover diverse acquisition conditions (e.g., varying b-values) and morphological variations, yielding robust prototypes tailored to the target clinical distribution.
Depending on the specific application, the class labels correspond to either the distortion status (e.g., for out-of-domain distortion assessment) or clinical quality scores (e.g., for PI-QUAL scoring).
These support samples are used to construct prototypes tailored to the target clinical distribution.
The prototype for each class is computed by averaging the feature embeddings of the $K$ support samples.
The remaining samples in $\mathcal{D}_{eva}$ constitute the query set $\mathcal{Q}_{eva}$. The probability of a query sample belonging to a specific class is predicted using the cosine similarity metric, same as $\mathcal{P}(y_j^q=c~|~x_j^q)$ defined in the meta-training phase.

\section{Experiments and Results}
\textbf{Implementation Details.}
Randomly initialized 3D ResNet-18 networks~\cite{hara2018can} are adopted as backbone encoders for DWI and T2WI feature extraction.
For meta-training, each episode is structured as a $2$-way, $5$-shot, $5$-query task. The weight in the adversarial loss $\lambda$ and the cosine similarity scaling $\tau$ are set to $0.5$ and $1$, respectively. The GRL scalar $\alpha$ is set to $1.0$, applying a linear warm-up from $0$ to $1.0$ during the first $10\%$ of training epochs.
The FiLM scaling and bias parameters are initialized as $\gamma(b) = 1$ and $\beta(b) = 0$.
We use the Adam optimizer with a learning rate of $1\times 10^{-4}$. The model is trained for $20$ epochs, where each epoch consists of $50$ episodes. All experiments are conducted on an NVIDIA Quadro GV100 GPU with 24 GB memory.

\begin{table*}[t]
\centering
\caption{Results of the performance comparison on the DWI distortion and PI-QUAL score assessment ($\geq 4$ vs. $< 4$) tasks. All methods were trained on the meta-training set from Dataset 1. All metrics are reported as mean with standard deviation (in brackets) over $10$ independent meta-training runs, evaluated using balanced accuracy (B-ACC)~\cite{brodersen2010balanced}, Sensitivity at $80\%$ Specificity (Sen@80Spe), and Specificity at $80\%$ Sensitivity (Spe@80Sen). The FSL baseline serves as the in-domain performance upper bound. The DWI distortion task is evaluated on both the in-domain meta-testing and out-of-domain PRIME datasets, while the PI-QUAL task is evaluated on the PRIME dataset. The bold font highlights the optimal performance among few-shot learning methods, and the asterisk $\ast$ denotes statistical significance ($p$-value $< 0.05$ by the one-tailed Wilcoxon signed-rank test) compared with the best competing few-shot method.}
\label{biao1}
\resizebox{\textwidth}{!}{
\begin{tabular}{l|ccc|ccc|ccc}
\hline
& \multicolumn{3}{c|}{DWI Distortion (Dataset 1)} 
& \multicolumn{3}{c|}{DWI Distortion (PRIME)}
& \multicolumn{3}{c}{PI-QUAL (PRIME)} \\
\cline{2-10}
\makecell{Method} & \makecell{B-ACC\\($\%$) $\uparrow$} & \makecell{Sen@80Spe\\($\%$) $\uparrow$} & \makecell{Spe@80Sen\\($\%$) $\uparrow$} 
& \makecell{B-ACC\\($\%$) $\uparrow$} & \makecell{Sen@80Spe\\($\%$) $\uparrow$} & \makecell{Spe@80Sen\\($\%$) $\uparrow$} 
& \makecell{B-ACC\\($\%$) $\uparrow$} & \makecell{Sen@80Spe\\($\%$) $\uparrow$} & \makecell{Spe@80Sen\\($\%$) $\uparrow$} \\
\hline
\makecell{PNet}   & \makecell{79.77\\(4.95)} & \makecell{79.90\\(4.55)} & \makecell{79.17\\(1.22)} 
& \makecell{56.42\\(4.21)} & \makecell{40.25\\(5.46)} & \makecell{36.80\\(3.89)} 
& \makecell{63.51\\(2.76)} & \makecell{49.28\\(2.50)} & \makecell{44.77\\(6.85)} \\
\makecell{SPNet}  & \makecell{84.97\\(0.69)} & \makecell{89.12\\(3.88)} & \makecell{\textbf{90.75}\\(4.69)} & \makecell{64.15\\(2.42)} & \makecell{52.66\\(4.26)} & \makecell{51.40\\(6.19)} 
& \makecell{65.65\\(1.82)} & \makecell{55.01\\(2.19)} & \makecell{47.70\\(0.96)} \\
\makecell{SGPL}   & \makecell{83.19\\(1.97)} & \makecell{85.66\\(6.15)} & \makecell{87.11\\(4.96)} 
& \makecell{62.81\\(1.94)} & \makecell{51.05\\(6.25)} & \makecell{48.95\\(4.52)} 
 & \makecell{65.39\\(0.44)} & \makecell{55.07\\(2.51)} & \makecell{46.45\\(0.72)} \\
\makecell{PNN}    & \makecell{83.95\\(0.44)} & \makecell{87.26\\(4.65)} & \makecell{88.54\\(6.50)} & \makecell{60.55\\(3.15)} & \makecell{48.30\\(4.56)} & \makecell{45.20\\(6.38)} & \makecell{63.56\\(3.45)} & \makecell{51.09\\(7.42)} & \makecell{42.99\\(6.15)} \\
\makecell{Ours}   & \makecell{\textbf{85.82}\\(1.31)} & \makecell{\textbf{93.39}$^{\ast}$\\(1.03)} & \makecell{89.90\\(1.51)} 
& \makecell{\textbf{69.75}\\(5.44)} & \makecell{\textbf{61.11}\\(5.91)} & \makecell{\textbf{61.10}\\(5.94)} & \makecell{\textbf{72.85}$^{\ast}$\\(3.17)} & \makecell{\textbf{65.22}$^{\ast}$\\(5.87)} & \makecell{\textbf{66.11}$^{\ast}$\\(3.43)} \\
\hline
\makecell{FSL}& \makecell{88.63\\(2.66)} & \makecell{97.02\\(1.70)} & \makecell{97.48\\(0.90)} & \makecell{-\\-} & \makecell{-\\-} & \makecell{-\\-} & \makecell{-\\-} & \makecell{-\\-} & \makecell{-\\-} \\
\makecell{FSL-5}  & \makecell{62.03\\(0.97)} & \makecell{38.06\\(2.70)} & \makecell{50.05\\(0.52)} & \makecell{51.84\\(1.93)} & \makecell{30.16\\(1.60)} & \makecell{17.19\\(3.43)} & \makecell{52.03\\(5.96)} & \makecell{30.88\\(2.51)} & \makecell{17.02\\(2.00)} \\
\hline
\end{tabular}}
\end{table*}

\textbf{Datasets and Data Preprocessing.}
A private MRI dataset~\cite{yan2024combiner} ('Dataset 1') is utilized for meta-training and meta-testing. The dataset consists of $1027$ valid high b-value ($\geq 1400$ s/mm$^2$) DWI volumes from $851$ patient cases. Each case is paired with a corresponding T2WI volume and a prostate gland mask annotated on the T2WI. We manually assessed the geometric consistency between the T2WI-based prostate mask and the imaging-coordinate-aligned DWI to obtain DWI distortion labels, yielding $790$ undistorted and $237$ distorted scans. Dataset 1 is randomly split into meta-training and meta-testing sets with an $8:2$ ratio at the patient level.
For out-of-domain and cross-task adaptation, we utilize the PRIME dataset~\cite{ng2025biparametric}, containing $483$ prostate MRI cases.
For the DWI distortion task, expert radiologists annotated $28$ cases as distorted and the remaining as undistorted. The dataset also includes PI-QUAL v1~\cite{giganti2020prostate} scores assigned by expert radiologists as part of the trial~\cite{ng2025biparametric}, where $31$ cases are low-quality scans (PI-QUAL $<$ 4) and the remaining are high-quality scans (PI-QUAL $\geq$ 4).
For preprocessing, DWI images are first aligned to the T2WI imaging coordinates, then resampled to a spacing of $[0.5, 0.5, 1.0]$ mm and center-cropped to $196\times196\times32$ using the SimpleITK library~\cite{lowekamp2013design}.


\begin{figure}[t] \centering \includegraphics[width=0.95\textwidth]{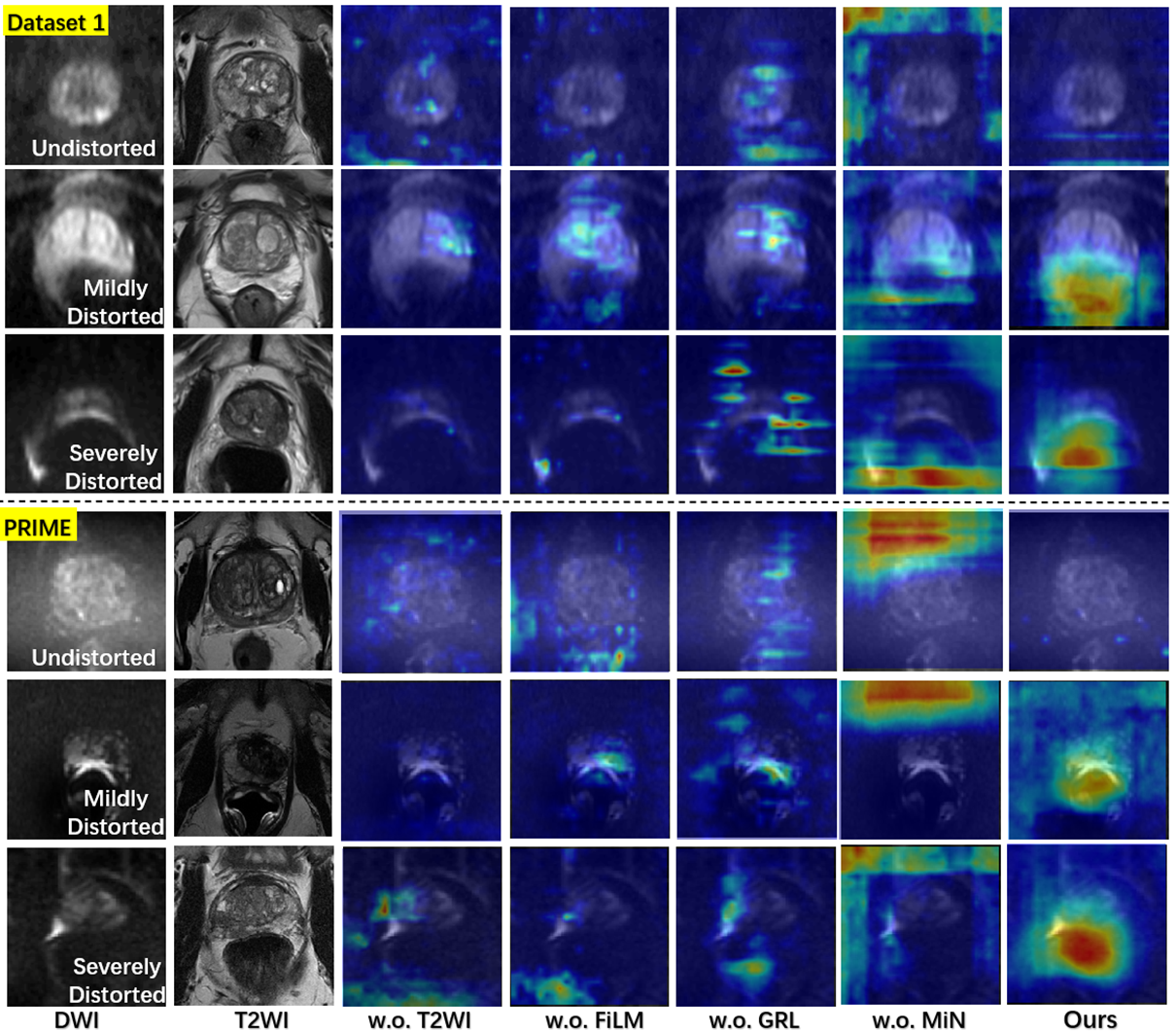} \caption{Grad-CAM heatmap visualizations of the ablation study on the meta-testing and adaptation set. 
Each row shows representative cases with undistorted, mildly distorted, and severely distorted DWI. Columns correspond to the full model (Ours) and its ablated variants, including without T2WI (w.o. T2WI), FiLM (w.o. FiLM), GRL (w.o. GRL), and MiN (w.o. MiN).} \label{tu2}
\end{figure}

\textbf{Comparison Study.}
We compare the proposed method with several prototypical few-shot learning baselines on the DWI distortion assessment task, including Prototypical Networks (PNet)~\cite{snell2017prototypical}, Siamese-Prototype Network (SPNet)~\cite{cheng2021spnet}, Semantic Guided Prototype Learning (SGPL)~\cite{li2025semantic}, and Prototype-Neighbor Networks (PNN)~\cite{jiang2025prototype}. All few-shot learning methods are trained on the meta-training set following the episodic meta-learning strategy described in Section 2.1. Adhering to the adaptation strategy in Section 2.2, evaluation involves two scenarios: in-domain adaptation on the meta-testing set and out-of-domain adaptation on the PRIME dataset.
Additionally, we include fully supervised learning baselines for reference: one (FSL) trained on the complete meta-training set to serve as an in-domain performance upper bound, and another (FSL-5) trained from scratch using only five support samples per class selected from the meta-testing set or the PRIME adaptation set. 
Due to the scarcity of distorted cases in the PRIME dataset, we only included the FSL-5 baseline for this dataset.
Unlike the episodic approach, these baselines are trained in a standard supervised manner using cross-entropy loss.
Table~\ref{biao1} shows that all methods perform reasonably well on the in-domain meta-testing set. The proposed method yields the best overall performance with a significantly superior Sen@80Spe, indicating an enhanced ability to detect distorted DWI images while maintaining high specificity.
On the out-of-domain PRIME adaptation set, the performance gap widens due to domain shift and severe class imbalance. Our method demonstrates significantly higher B-ACC, Sen@80Spe and Spe@80Sen than all baselines.
This demonstrates its robust cross-dataset generalization under few-shot setting, with reduced risks of both missed detections and false rejections, which is a critical advantage for clinical deployment.

\textbf{Ablation Study.}
We performed an ablation study to quantify the contribution of key components: the T2WI branch, FiLM, GRL, and MiN modules. All models were trained and evaluated under the same settings as in Table~\ref{biao1}.
We visualize the Grad-CAM attention maps~\cite{selvaraju2017grad} on both the meta-testing and the out-of-domain PRIME dataset, as shown in Fig.~\ref{tu2}. The models without MiN exhibit severe attention leakage into the background air and surrounding extra-prostatic tissues, such as the rectum (e.g., the high activation in the anterior or posterior regions), confirming that MiN effectively suppresses irrelevant background noise which could otherwise be used for prediction probably due to the lack of minor class samples. Without the T2WI branch, the model lacks anatomical reference, leading to inaccurate distortion localization in the attention maps.
On the other hand, removing the b-value alignment modules (FiLM and GRL) results in diffuse or misaligned activation patterns.
In particular, the heatmaps for w.o. FiLM and w.o. GRL mainly appear blue with lower activation intensities, indicating reduced prediction confidence in identifying distortion features under varying b-values. In contrast, our method successfully highlights the distortion areas in low-quality scans, aligning well with the ground-truth anatomical deformations.
Quantitatively, for Sen@80Spe and Spe@80Sen, the FiLM and GRL modules yield the most significant improvements, with average gains ranging from $14.48\%$ to $19.64\%$. The T2WI branch contributes a performance boost of $10.63\%-11.63\%$, while the MiN module yields improvements of $6.84\%-9.17\%$.
Both visualization and quantitative results confirm that all components contribute to performance gains and are essential.

\textbf{Application on PI-QUAL Scoring System.}
Finally, we adapt the trained model to the PI-QUAL score assessment task on the PRIME dataset, aiming to classify whether the PI-QUAL score is $\geq 4$. The models in the comparison study are used, which were trained exclusively on the DWI distortion task without PI-QUAL score labels. 
Due to the scarcity of low PI-QUAL cases in the PRIME dataset, we only included an FSL-5 baseline trained from scratch on five PRIME support samples per class.
As shown in Table~\ref{biao1}, the proposed method shows the strongest cross-task and cross-dataset adaptation performance among all baselines. It outperforms all baselines on B-ACC, Sen@80Spe and Spe@80Sen, with improvements ranging from $10.15\%$ to $21.12\%$. The superior performance in Sen@80Spe and Spe@80Sen demonstrates that our method effectively balances the risks of missing poor-quality images and rejecting usable ones. Furthermore, the poor performance of the supervised baselines (FSL-5) highlights the difficulty of this task under severely class-imbalanced distribution and limited annotation. The robust performance of our method suggests that the distortion-sensitive features learned from the DWI distortion task transfer effectively to the PI-QUAL score assessment task on PRIME, despite the dual-scarcity constraint.

\section{Conclusion}
We propose a few-shot biparametric prototypical network to address data scarcity and domain shifts in automated prostate MRI quality assessment. Meta-trained purely on distortion data, our model adapts to out-of-domain datasets to predict complex PI-QUAL scores using just five representative samples per class, significantly outperforming baselines. While inherently sensitive to support set selection, we mitigate this by choosing diverse samples that act as expert-defined "golden standard" templates, directly mirroring clinical workflows. Ultimately, this robust, data-efficient paradigm shows broad potential for distinguishing optimal (PI-QUAL 5) from sub-optimal ($\leq 4$) scans and facilitating reliable downstream applications like automated PI-RADS assessment~\cite{tang2025impact}.

\subsubsection{Disclosure of Interests.} The authors have no competing interests to declare that are relevant to the content of this article.

\bibliographystyle{splncs04}
\bibliography{bib.bib}
\end{document}